\documentclass{article}

\usepackage{arxiv}

\usepackage[utf8]{inputenc} 
\usepackage[T1]{fontenc}    
\usepackage{hyperref}       
\usepackage{url}            
\usepackage{booktabs}       
\usepackage{amsfonts}       
\usepackage{nicefrac}       
\usepackage{microtype}      
\usepackage{lipsum}
\usepackage{graphicx}
\graphicspath{ {./images/} }
\usepackage{float}

\title{\textbf{Me}dAI \textbf{Di}alog \textbf{C}orpus (MEDIC): Zero-Shot Classification of Doctor and AI Responses in Health Consultations}

\author{
Olumide E. Ojo \\
Instituto Politécnico Nacional (IPN), \\
Centro de Investigación en Computación (CIC), \\
Mexico City, Mexico.\\
\texttt{olumideoea@gmail.com} \\
   \And
Olaronke O. Adebanji\\
Instituto Politécnico Nacional (IPN), \\
Centro de Investigación en Computación (CIC), \\
Mexico City, Mexico.\\
\texttt{olaronke.oluwayemisi@gmail.com} \\
\And
Alexander Gelbukh \\
Instituto Politécnico Nacional (IPN), \\
Centro de Investigación en Computación (CIC), \\
Mexico City, Mexico.\\
\texttt{gelbukh@cic.ipn.mx} \\
   \And
Hiram Calvo\\
Instituto Politécnico Nacional (IPN), \\
Centro de Investigación en Computación (CIC), \\
Mexico City, Mexico.\\
\texttt{hcalvo@cic.ipn.mx} \\
  \And
Anna Feldman\\
Montclair State University, \\
Montclair, USA.\\
\texttt{feldmana@montclair.edu} \\
}

\begin{document}
\maketitle
\begin{abstract}
Zero-shot classification enables text to be classified into classes not seen during training. In this study, we examine the efficacy of zero-shot learning models in classifying healthcare consultation responses from Doctors and AI systems. The models evaluated include BART, BERT, XLM, XLM-R and DistilBERT. The models were tested on three different datasets based on a binary and multi-label analysis to identify the origins of text in health consultations without any prior corpus training. According to our findings, the zero-shot language models show a good understanding of language generally, but has limitations when trying to classify doctor and AI responses to healthcare consultations. This research provides a foundation for future research in the field of medical text classification by informing the development of more accurate methods of classifying text written by Doctors and AI systems in health consultations.
\end{abstract}


\section{Introduction}
Providing, maintaining, and restoring health is an integral part of healthcare~\cite{snelling2014introduction}. Individuals, families, and communities depend on healthcare for their physical, mental, and social needs. There are different health systems (HS) in different countries, but they all strive to provide high-quality, affordable healthcare for all. There is a wide range of medical professionals (MPs) in the healthcare industry (HI) who collaborate to provide comprehensive health care~\cite{mitchell2008multidisciplinary}. Doctors, nurses, therapists, dentists, pharmacists, and others are included in this group of MPs. Health professionals also offer medical education and disease prevention to patients. Having good health is essential not only for our well-being, but also for the progress of our society. A healthy population is more productive, more educated, and capable of contributing to society's progress and stability.

In recent decades, the HI has faced a number of challenges and undergone significant transformations~\cite{dal2023challenges}. This sector faces several challenges, including rising costs, new disease epidemics, a limited number of professionals, and the necessity of adapting to changing needs. As healthcare continues to evolve, organized HS, the expansion of ME, health insurance schemes, and the use of artificial intelligence (AI) systems have accompanied it~\cite{siala2022shifting}. Medical knowledge, technology, and changing needs of individuals and societies have contributed to advances in healthcare practices (HP), enabling patients to interact and ask questions more effectively and efficiently. The integration of technology in HP has revolutionized the way patients seek information and assistance. With search engines, social media platforms, online forums, telemedicine platforms, and chatbots, patients can ask questions and receive automated responses from humans and machine learning (ML) models alike. However, when it comes to the proper diagnosis of the patient, these technologies have inherent limitations that must be addressed. Despite the fact that Doctors provide objective and formal responses, machines and models often struggle to understand patients' emotions, subjective experiences, and physical health statuses~\cite{adams2017dehumanizing, yun2021behavioral}. These details are crucial in health diagnosis (HD), as they provide valuable information that can aid in accurate treatment. Doctors, who adhere to established rules, protocols, and professional standards, provide patients with the best medical care based on factual information, evidence-based analysis, and an unbiased approach.

Distinguishing between text authored by humans and text generated by AI models is an important part of the NLP field. Text classification encompasses a wider range of classification tasks~\cite{ojo2022language, adebanji2022sequential, 9677627, prabhakar2021medical, 9451752, kumar2022deep, LI2021345, liao2023differentiate, ojo2023legend, ojo2020sentiment, ojo2021performance,  ojo2023transformer, srivastava2020healthcare, ojo2022automatic, 9185650}. Human authors are influenced by their preconceived idea and unique perspectives, resulting in text that exhibits nuances, affection, and variability in writing styles. AI-generated text, on the other hand, which is produced by automated systems, ML models, or chatbots, uses algorithms and statistical patterns to generate text that mimics human language. Classifying text of this kind requires advanced methods that analyze semantic meaning, contextual understanding, and machine-generated text-specific language patterns. Linguistic and NLP techniques can help distinguish between human-generated and AI-generated text. In order to protect human life, promote transparency, and maintain the highest ethical standards in the HI, accurate text attribution is of utmost importance. To make informed decisions that directly affect patient care and well-being, it is necessary to communicate accurately, obtain reliable information, and establish trust, which may not be guaranteed when interacting with AI models.

The advancement of NLP algorithms and models is highly dependent on the availability of text corpora for training, evaluation, and development~\cite{mitrovic2023chatgpt, Faqar-Uz-Zamane041396, melvin-etal-2004-creation}. Creating a dataset to help distinguish between doctor and AI responses to patient questions is of the utmost importance in ensuring accurate diagnosis and appropriate treatment plans. Models can learn to accurately detect text from human and AI sources and help patients build trust and make informed decisions about their health. A patient should be able to determine whether they are receiving responses from a qualified healthcare professional or an automated system. By identifying human responses, patients can rely on the expertise of the MPs to receive accurate diagnoses and effective treatment options based on their unique medical conditions. In addition, the differentiation between doctor and AI responses contributes to building trust in the healthcare sector. Knowing who provides medical advice can give patients and the community confidence in the integrity and reliability of the healthcare system. Improved patient experiences and outcomes can be achieved by providing transparency and clarity on the origin of text received by patients.

In ~\cite{mitrovic2023chatgpt}, the authors investigated whether ML models can effectively differentiate between human-generated and machine-generated text, specifically in short online reviews. As part of their research, researchers attempted to determine why the model made this decision, while distinguishing between text that was generated by machines and text that was generated by humans. Two experiments were carried out, one involving ChatGPT-generated text from custom queries and the other involving rephrased human-generated reviews. For our investigation conducted in the HI, we developed a comprehensive dataset consisting of text from three sources. The dataset included original doctor responses to patient questions, ChatGPT-generated text from custom queries of the same questions, and rephrased versions of the original doctor responses. Analysis of our dataset provided valuable information on the writing style of ChatGPT-generated text in medical responses. However, certain characteristics were identified in the ChatGPT-generated text that distinguished it from the original doctor's responses. Further analysis revealed that the ChatGPT-generated text consistently emphasized the importance of consulting suitable MPs or visiting a hospital for HD. Although this guide aligns with responsible healthcare advice, it also highlights the limitations of the AI system in providing comprehensive medical diagnoses or treatments. ChatGPT's responses acknowledged the need for direct interaction with medical experts to ensure appropriate and accurate care.

\emph{Zero-shot classification~\cite{veeranna2016using} has emerged as a widely accepted approach in various NLP tasks that lack sufficient training data. We use this approach in particular to analyze the need for specific medical corpus training for the classification of responses from health consultations into doctor- and AI-generated text. As AI-generated responses in the medical field are constantly evolving, will zero-shot classification be a solution that leverages transfer learning to solve the problem of detecting doctor- and AI-generated text in health conversations?} In zero-shot classification, a pre-trained language model is used as a starting point. These models are trained on a vast corpus of text from the Internet and have learned to capture a rich understanding of language. During training, the model learns to predict the next word in a sentence given the preceding context. This pre-training enables the model to develop a strong language understanding that can be fine-tuned for many tasks. In this classification task, the pre-trained model is provided with a description of each class it needs to classify. The model then learns to map the text input to the appropriate class based on its understanding of the relationships between words and concepts.  

\begin{figure*}[!ht]
  \centering
  \includegraphics[width=0.8\textwidth]{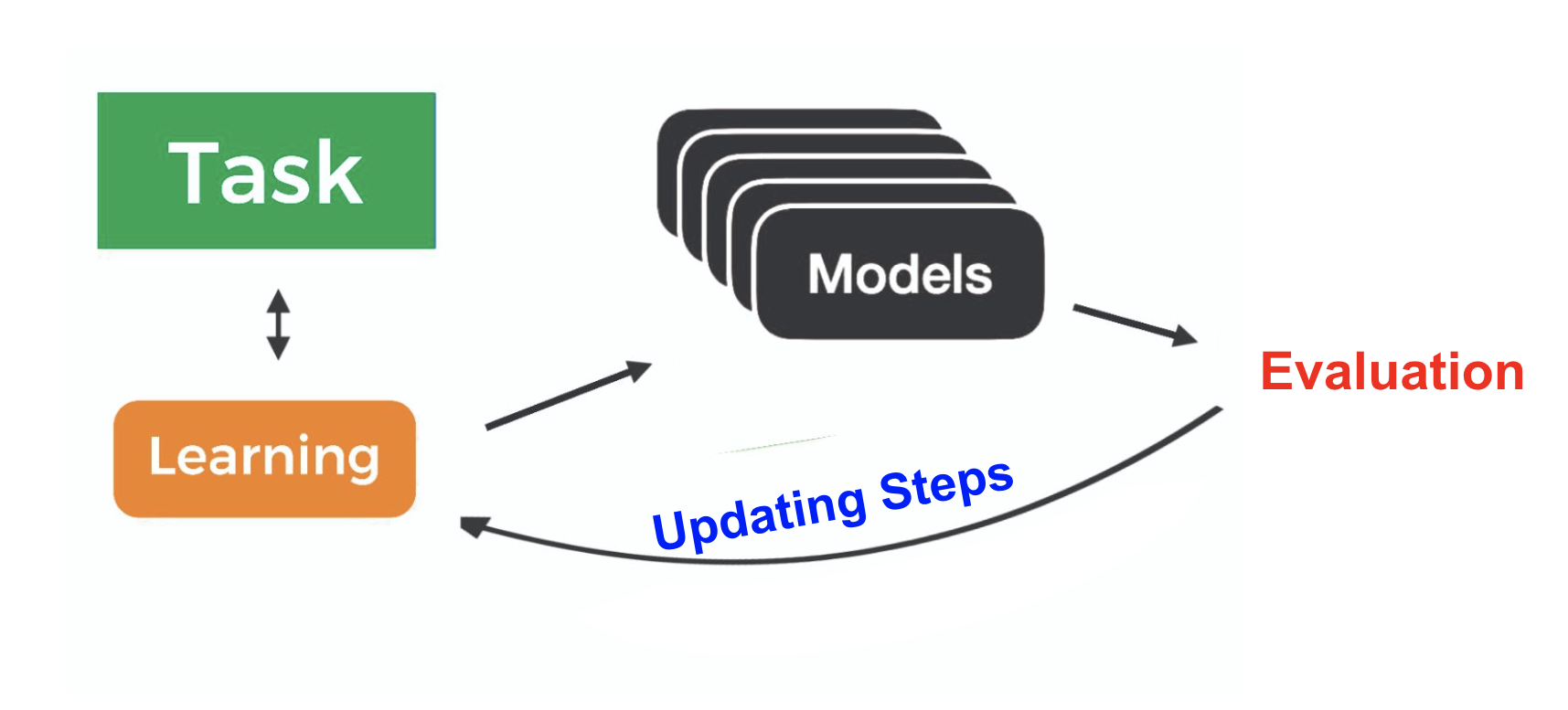}
  \caption{Architecture of the Zero-Shot Classification Experiments}
  \label{fig:fig6}
\end{figure*}

Understanding the differences between AI-generated text and original doctor responses is essential to effectively leverage AI systems in healthcare settings and to ensure that patients receive the most appropriate and reliable ME and guidance. In the rephrased doctor's response, medical terminologies and vocabularies were seamlessly replaced, producing text that closely resembled human-generated text. This finding poses a significant challenge in accurately detecting the source of the text. Addressing this concern requires the development of robust and sophisticated methods for source attribution and contextual comprehension. By fine-tuning ML models and improving model training, we can detect and classify machine-generated responses, and patients can benefit from the distinctive expertise and empathy that human-doctor's provide.~\emph{To establish a baseline for our task, we evaluated the capability of pre-trained language models to perform zero-shot classification effectively and identify the sources of text within the corpus}. This evaluation will be conducted through three distinct experiments, without the use of specific corpus training.

The contributions of this paper are summarized as follows:

\begin{itemize}
  \item A dataset comprising responses from Doctors and AI to patient inquiries.
  \item We offer insights into patterns, distribution, and relationships in data for researchers.
 \item We explore pre-trained models for zero-shot classification to identify AI-generated and human-generated responses to patient questions.
 \item We analyzed the dataset as binary and multiclass tasks for human-AI text classification.
\end{itemize}

This study consists of three subtasks that aim to distinguish between doctor-generated healthcare consultations and AI-generated ones.

\begin{itemize}
  \item First subtask - This centered on distinguishing between ChatGPT-generated text and the responses provided by doctors. 
  \item Second subtask - This involved differentiating between the rephrased text from the doctors' response and the actual responses provided by the doctors.
  \item Third subtask: This involved combining doctor responses, ChatGPT-generated responses, and rephrased responses and detecting each in a multi-class analysis.
\end{itemize}

The remainder of the paper is organized as follows. Section 2 presents the review of the related literature, Section 3 discusses the methodology used to create the corpus, and Section 4 presents the analysis of the corpus. In Section 5, we describe the experimental setup and present the results of the three subtasks carried out in this study. Section 6 discusses the findings of this study, addressing the implications and potential applications of this task in the healthcare sector. Finally, Section 7 concludes the paper with a summary of the key contributions, limitations, and future directions for research in this field.

\section{Literature Review}
AI-powered systems, also known as machines, have become valuable tools to support healthcare professionals by providing accurate and timely responses to patient queries and concerns. These systems use AI algorithms to understand and address a wide range of health-related questions, and healthcare professionals can now improve their efficiency during healthcare delivery. Doctors, on the other hand, undergo extensive training and have years of clinical experience, which allows them to provide accurate diagnoses and treatment recommendations. It is necessary to ensure that the responses delivered by these AI models are certified by medical experts before they are presented to patients. AI-generated responses are inadequate for diagnosing patients and must be identified and classified.

Based on research by Baker et al.~\cite{baker2020comparison}, virtual AI assistants have been used to support overwhelmed healthcare systems by offering medical advice to patients. To gauge their efficacy, researchers compared the accuracy and safety of AI systems with human-Doctors using the same set of clinical cases. Data was collected from both AI and human-Doctors, revealing that the AI system could diagnose patients with a level of accuracy and safety comparable to that of human physicians. This direct comparison underscores the reliability of AI-powered systems, especially considering the inconsistencies often found in the diagnostic decisions of human physicians.

\cite{shoham2023cpllm} introduced a method called Clinical Prediction with Large Language Models (CPLLM) for the prediction of clinical diseases. Their experiment involved fine-tuning a pre-trained LLM using prompts and historical diagnosis records to predict whether patients will be diagnosed with a specific disease in their next visit or subsequent diagnosis. The study compares CPLLM to various baselines, including Logistic Regression, RETAIN, and Med-BERT, a current state-of-the-art model for disease prediction using structured Electronic Health Record (EHR) data. The experiments demonstrate that CPLLM outperforms all tested models in terms of both PR-AUC and ROC-AUC metrics and provides significant improvements over baseline models. CPLL performs well with diverse datasets and handles different code-types and data-types. CPLLM presents a novel and effective approach to clinical disease prediction based on patients' clinical history, outperforming current state-of-the-art methods.

In a study by Faqar et al.~\cite{Faqar-Uz-Zamane041396}, researchers examined the diagnostic precision and influence of Ada, an app-based diagnostic instrument, on patient outcomes. The study primarily gauged Ada's diagnostic accuracy by comparing its suggested diagnoses with the conclusive diagnosis at discharge. Ada's diagnostic precision and treatment timing were juxtaposed with conventional doctor-patient consultations. Furthermore, the research explored other factors such as complication rates, duration of hospital stay, and patient survival rates. This investigation of Ada's diagnostic influence in emergency room scenarios offers insights into the efficiency and potential advantages of app-based diagnostic tools.

The authors in~\cite{melvin-etal-2004-creation} presented a comprehensive analysis of a doctor-patient dialogue corpus developed for English-Persian speech-to-speech machine translation in medical contexts. The corpus creation process involved recording and transcribing dialogues between medical students and standardized patients, which were later translated into Persian. The authors discussed the advantages and disadvantages of this approach. Benefits include the ability to tailor the corpus according to specific needs, which would be impractical with real doctor-patient data, as well as the avoidance of privacy and legal issues. The paper effectively addressed concerns surrounding the authenticity of dialogues and emphasizes the significance of such data for the development of systems in the medial field. The outcome of their proposed method was successful in generating medical interaction data, providing a solution to the privacy and legal obstacles associated with genuine medical dialogues.

In a thorough review by Dave et al.~\cite{dave2023chatgpt}, the prowess of ChatGPT, a sophisticated language model powered by deep learning, was examined. The authors highlighted ChatGPT's impressive capability to understand and produce context-relevant replies across a spectrum of prompts. They also illuminated its potential roles in the medical sector, encompassing research, diagnosis, education, and patient care. However, the review critically examines the ethical and legal challenges related to ChatGPT, including issues such as copyright breaches, potential medical legalities, and the imperative for transparency in AI-generated outputs. Despite its potential, the authors stressed the need for a discerning approach to ChatGPT's limitations and ethical implications. They recommend further studies to address both these challenges and tap into ChatGPT's potential benefits in healthcare.

In the study by~\cite{9185650}, the researchers introduced a rule-based strategy for medical text classification, using easily understandable regular expressions. These expressions were automatically crafted using a constructive heuristic technique and further refined with pool-based Simulated Annealing (PSA). Recognizing the issues of interpretability in deep neural network (DNN) strategies, the authors emphasized the importance of their rule-based method in the medical domain. Their approach efficiently crafts regular expressions for large datasets, ensuring top-tier results. The introduced PSA method autonomously refines machine-derived regular expressions. This approach, when tested with real-world data from a leading Chinese online medical platform, outperformed other meta-heuristic techniques, such as Genetic Programming. The study concludes that its methodology complements existing ML strategies, offering clear and understandable results in text classification tasks.

Zero-shot classification was proposed in~\cite{lupart2022zero} as a suitable approach to timely categorization of articles with MeSH (Medical Subject Headings) categories, a comprehensive thesaurus to index biomedical publications. The study investigates the integration of MeSH semantic information to enhance BioBERT representations for zero-shot/few-shot tasks. To achieve this, a multitask learning approach was used, incorporating a seq2seq task to induce the MeSH hierarchy within the representations. The experimental results, conducted on the MedLine and LitCovid datasets, demonstrated that the resulting representations effectively capture the hierarchical relationships present in MeSH. Although the multitask framework does not yield significant performance improvements in zero-shot and few-shot scenarios, positive outcomes are observed in precision and structural probing tasks, indicating the model's effective capture of semantics.

In the research presented by ~\cite{mascio2020comparative}, the authors delved into various text classification techniques, specifically designed to interpret electronic health records. The study grappled with the intricacies of medical terminology and the distinctive language patterns seen in clinical documentation. The researchers scrutinized the effects of various word representations, text preprocessing methods, and classification algorithms in four different text classification tasks. Interestingly, the findings indicated that custom-tailored traditional techniques, finely tuned to the linguistic texture and architecture of clinical text, could match or even surpass the efficacy of newer context-based embedding methods like BERT. The study cast a wide net, encompassing different word representation strategies, ranging from the bag-of-words model to traditional embeddings grounded in both specific and generic datasets and up to contextual embeddings. Additionally, they probed various text pre-processing and tokenization techniques. The data revealed a noteworthy insight: while contextual embeddings shine in situations resistant to customization, Bi-LSTM—when married to precise entity extraction tasks and domain-focused embeddings—tends to overshadow these embeddings. The conclusions drawn from this study illuminate the nuanced performance of different methodologies in the classification of medical texts, serving as a compass for subsequent research and real-world implementations.

The study conducted by \cite{liu2023deid} addresses the critical need for effective and efficient de-identification techniques within the healthcare sector to maintain confidentiality and privacy. By examining the limitations of current methods, the authors explore the potential of large-language models (LLMs) such as ChatGPT and GPT-4. They introduce a novel de-identification framework, DeID-GPT, which leverages the capabilities of LLMs to identify and redact identifying information from unstructured medical text. The results indicate that DeID-GPT surpasses existing methods, offering enhanced accuracy and reliability in preserving the original text structure and meaning while effectively masking private information. This research makes a significant contribution by pioneering the application of ChatGPT and GPT-4 in medical text data processing and de-identification, providing valuable insights for future research and solution development in the healthcare domain.

To process data from health-related social media,~\cite{9451752} highlighted the use of supervised learning algorithms as a preferred method for optimal performance. Deep learning techniques were examined, including Convolutional Neural Networks and Recurrent Neural Networks, which have gained popularity in the healthcare sector due to their ability to handle complex data. The models explored offer an effective and efficient analysis of large health datasets, enabling the discovery of meaningful patterns that traditional analytics struggle to identify. The authors noted that deep learning models excel in recognizing patterns in social health networks. Their work focused on investigating models applied to text classification in social media healthcare networks and tried to improve the performance of text classifiers by employing suitable methodologies, with the goal of achieving promising results in future applications.

The potential of ML and automated data processing in the biomedical and public health domains was explored in ~\cite{8733799}, focusing on the use of social media data. The article identified that consumer health terminology was rarely incorporated into social media text analysis. To address this, the authors introduced the Medical Social Media Text Classification Algorithm (MSMTC), which integrates consumer health terminology. The algorithm comprises two main tasks: extraction of consumer health terminology and text classification. The approach used a double channel structure for training and employed an adversarial network to extract consumer health terminology. The extracted terminology is then used for text classification using a double channel subtraction method. The proposed algorithm was evaluated using datasets that contain patient descriptions on social media. The result demonstrated that the MSMTC algorithm outperforms other methods and baseline models. 

In their article, ``Classification of Medical Sensitive Data based on Text Classification"~\cite{8991726}, the authors tackled the issue of data sensitivity in the rapidly expanding volume of patient privacy data. They proposed the application of text classification techniques, as described in their work~\cite{8991726}, to address the issue of classification of sensitive medical information. With the widespread adoption of electronic storage for medical records, ensuring the security and confidentiality of these data has become increasingly critical. The authors recognized this challenge and provided a solution using text classification methods. They conducted an experiment to test the accuracy of classifying sensitive data within medical records and successfully demonstrated the effectiveness and feasibility of their approach.

In their study, Prabhakar and Won~\cite{prabhakar2021medical} investigated the application of automatic medical text classification in the field of healthcare. The research addressed the challenge of extracting valuable information from clinical descriptions by proposing two innovative deep learning architectures: the quad channel hybrid long- and short-term memory model (QC-LSTM) and the hybrid bidirectional gated recurrent unit (BiGRU) model with multi-head attention. These approaches were able to minimize the human effort required to label training data and effectively processed large volumes of detailed patient information. The proposed models were applied to two distinct medical text datasets and their performances were evaluated. The findings underscore the importance of automated clinical text classification to unlock structured information embedded within the clinical text, including disease specifications and related pathological conditions.

The approach outlined in \cite{9677627} was formulated to tackle the challenge of imbalanced datasets. Their strategy involved integrating short text sequences into the training process to enhance the model's accuracy in handling infrequent classes. The methodology encompassed the compilation of a collection of keywords, including concise phrases related to each class. Employing a CNN model for text classification, the study established that incorporating keywords and brief phrases for each class within the training dataset substantially heightened the model's efficacy in addressing rare classes. These keywords were integrated as additional data during each training batch, contributing to a training loss that amalgamates input from both original data and keywords. The effectiveness of this approach was evaluated in the context of classifying cancer pathology reports. The outcomes showcased a marked enhancement in the model's performance with regard to infrequent classes. The influence of keywords on model output presented a means of surmounting model limitations when faced with scenarios characterized by limited training data.

\cite{liao2023differentiate} discussed the responsible and ethical use of artificial intelligence-generated content (AIGC) in medicine. The authors developed ML methods to effectively identify the source of medical texts. The researchers compiled datasets of human-written medical texts and ChatGPT-generated texts, analyzing various linguistic features such as vocabulary, part-of-speech, dependency, sentiment, and perplexity. They find that human-written texts are more detailed and diverse, while ChatGPT-generated texts prioritize fluency and logic, but lack context-specific information. To detect ChatGPT-generated medical texts, the researchers employed a BERT-based model and achieved outstanding performance. The study emphasized the importance of recognizing and addressing linguistic differences between human- and AI-generated medical content to ensure responsible use of large language models in medicine. The authors stress the need for careful implementation and an open discussion on the benefits and risks of AI in the medical field. They highlight the potential harm caused by inaccurate information generated by AI systems like ChatGPT and provide a demonstration for identifying such content. 

The research by ~\cite{srivastava2020healthcare} addressed the challenge of improving the accuracy of ML-based text classification systems. The study identified the limitation of existing systems in effectively modeling input text data and its noise characteristics, leading to low accuracy. To overcome this limitation, the authors proposed a novel concept called misrepresentation ratio (MRR) as a metric to evaluate the ML system's performance. The MRR incorporated attributes such as data size, classifier type, partitioning protocol, and percentage of misrepresentation to assess system effectiveness. The research used a comprehensive data analysis using diverse text datasets and various classifiers with different training protocols. The ML system achieves significant improvements in classification accuracy. In particular, an MLP-based neural network achieved the highest accuracy, demonstrating a 6\% improvement compared to previous studies. The decrease in MRR and the increased stability of the system further validate the robustness of the system. Their research introduced the innovative concept of the misrepresentation ratio and showcased its effectiveness in improving accuracy and robustness in ML-based text classification. 

In this research, we reviewed numerous studies related to AI in healthcare and NLP. Developing accurate classification models that can identify text generated by humans and AI is becoming increasingly important. Zero-shot classification in healthcare has demonstrated potential benefits and challenges. The development of classification models for healthcare should be supervised and informed by domain-specific knowledge. Based on our analysis, we have identified gaps that need to be addressed and aim to contribute to the field by proposing new methodologies for accurate classification of doctor and AI responses.

\section{Corpus Creation}
QuestionDoctors (QD)~\cite{QuestionDoctors} is an online platform used to connect patients with medical professionals for answers to health-related questions. By being able to access medical information and guidance from the comfort of their own home, users can seek help without the need to make a physical appointment. One of the key features of QD is the ability to ask questions directly to medical professionals. The question submission process is straightforward and requires users to provide relevant details about their concern, such as symptoms, medical history, and any additional information that may help diagnose their problems. The platform emphasizes the importance of accurate and complete information to facilitate accurate responses. Once a question is submitted, users can expect to receive responses from certified doctors within a reasonable time frame. The quality and thoroughness of the responses depend in large part on the complexity of the question and the availability of Doctors specializing in the relevant field. 

The medical question/answer dataset was obtained from the QD website as a valuable resource for researchers and developers in the field of healthcare and NLP. The dataset is accessible through the GitHub link~\cite{github}. The repository includes medical questions and answers collected from various websites. For this analysis, the QD website dataset was used. The questions in the dataset were posed to ChatGPT~\cite{chatgpt} and the response provided by the doctors was rephrased. RapidAPI offers APIs for a wide range of purposes, including data retrieval, paraphrasing, and access to a wide range of functionality. To paraphrase the text, we used the RapidAPI Paraphrasing and Rewriter API~\cite{rapid} with the appropriate endpoint URL and API key.  We defined two functions: one for paraphrasing text via the API and another for handling large texts that exceed the API's limits. The core activity in this task involves sending the text to the API and receiving a paraphrased version. When the text is too lengthy, it splits into smaller chunks, each processed separately. The script then stores the results back into a file, while also dealing with any error that might come up with network requests or data processing. 



The answers provided by ChatGPT and the rephrasing of the Doctors' response were recorded and used to classify text generated by AI and those generated by humans in healthcare consultations. When comparing Doctor's answers with AI-generated answers, insights were gained into the strengths and limitations of the AI model in understanding medical inquiries. Researchers can train and fine-tune models using this dataset to improve their ability to classify AI-generated responses in healthcare settings. When comparing AI-generated answers with those provided by Doctors, it becomes possible to assess the suitability and relevance of the models' output. This analysis is crucial to evaluate the performance of AI systems and understand their potential impact on healthcare delivery. It can help identify areas where AI-generated responses excel or fall short, informing future advances in technology. Understanding the capabilities and limitations of AI in healthcare is essential for the responsible integration and deployment of these systems. When working with this dataset, it is important to consider the context in which the questions were posed and the rephrased responses provided. 

Table~\ref{tab:tab7} presents an overview of the datasets. As shown in the table, the Doctor-ChatGPT (DC) dataset contains 3760 instances divided into 2 classes. Again, the Doctor-Rephrased Doctor (DR) dataset also contains 3760 instances divided into 2 classes. In addition, the Doctor-ChatGPT-Rephrased Doctor (DCR) dataset contains 5640 instances divided into 3 classes, which combines the interactions from the first two datasets. This data set will allow for a complex analysis compared to the first two datasets.

\begin{table*}[!ht]
  \centering
  \small
  \caption{Statistics of the Dataset}
  \label{tab:tab7}
  \begin{tabular}{c|c|c}
    \hline\hline
    Dataset & Number of Classes & Size \\
    \hline\hline
    DC & 2 & 3760\\
    \hline
    DR & 2 & 3760\\
    \hline
    DCR & 3 & 5640\\
    \hline\hline
  \end{tabular}
\end{table*}

Tables~\ref{tab:tab1}-~\ref{tab:tab3} present a representative sample of the dataset, illustrating the question posed, the corresponding response from a doctor, the rephrased response, and the response generated by ChatGPT. Table~\ref{tab:tab1} shows some questions asked in healthcare consultations, along with the corresponding response from a doctor. The doctor provided helpful responses, addressing the specific concerns raised by the patients. 

\begin{table*}[!ht]
  \centering
  \small
  \caption{Questions and Doctor's Responses in Healthcare Consultations}
  \label{tab:tab1}
  \begin{tabular}{p{0.35\linewidth}|p{0.55\linewidth}}
    \hline\hline
    Question & Doctor's Response \\
    \hline\hline
    Is my Anti HIV Test Conclusive after 241 days or need retest? & Your time slots of the tests and the final result even after 241 days is indicative of HIV-free status. So, just relax. And to avoid such anxiety and fear in the future, just avoid multiple partners. Hope this helps.\\
    \hline
    I developed breathing difficulties approx. 2 years ago, which have worsened daily since. What can I do? & In my opinion, if you have something other than anxiety you should check thyroid glands and hiatal herniation possibilities and then i recommend yoga breathing exercises and easiness techniques.\\
    \hline
    Low resting heart rate. Can it be secondary adrenal insufficiency? If not, what is the cause? & This is normal after severe stressful traumatic events. No need to worry now, just try to calm down for some time and it will resolve on its own. If it persists, then cortisol hormone thyroid profile will be needed.\\
    \hline\hline
  \end{tabular}
\end{table*}

In Table~\ref{tab:tab2}, the same set of questions was posed to ChatGPT and its corresponding responses were recorded. ChatGPT offers informative responses, providing general guidance and suggestions based on the given information. ChatGPT emphasizes the importance of seeking professional medical help, as seen in the most commonly used words depicted in Figure~\ref{fig:fig1}.

\begin{table*}[!ht]
  \centering
  \small
  \caption{Questions and ChatGPT's Responses in Healthcare Consultations}
  \label{tab:tab2}
  \begin{tabular}{p{0.35\linewidth}|p{0.55\linewidth}}
    \hline\hline
    Question & ChatGPT's Response \\
    \hline\hline
    Is my Anti HIV Test Conclusive after 241 days or need retest? & An anti-HIV test conducted after 241 days (approximately 8 months) is generally considered conclusive. The window period for most HIV tests is around 3 months, and a negative result at 8 months is highly reliable. However, it's always a good idea to consult with a healthcare professional or a testing center to confirm whether a retest is necessary based on your specific circumstances.\\
    \hline
    I developed breathing difficulties approx. 2 years ago, which have worsened daily since. What can I do? & Experiencing worsening breathing difficulties over a period of two years is concerning. It is crucial to seek immediate medical attention and consult a healthcare professional, preferably a pulmonologist or respiratory specialist. They can conduct a thorough evaluation, order appropriate tests, and provide a diagnosis and treatment plan to address your worsening symptoms. Do not delay in seeking medical help for your condition.\\
    \hline
    Low resting heart rate. Can it be secondary adrenal insufficiency? If not, what is the cause? & A low resting heart rate can have various causes, and secondary adrenal insufficiency can be one possible cause. However, it is important to consult with a healthcare professional for a comprehensive evaluation to determine the underlying cause of your low resting heart rate. Other potential causes may include medication side effects, certain medical conditions, or physiological factors. A healthcare professional can assess your specific situation and provide an accurate diagnosis and appropriate treatment options.\\
    \hline\hline
  \end{tabular}
\end{table*}

Table~\ref{tab:tab3} shows the rephrased responses provided by the doctor. By rephrasing, we can understand different communication styles and variations in machine interactions within healthcare delivery settings.

\begin{table*}[!ht]
  \centering
  \small
  \caption{Questions and Rephrased Doctor's Responses in Healthcare Consultations}
  \label{tab:tab3}
  \begin{tabular}{p{0.35\linewidth}|p{0.55\linewidth}}
    \hline\hline
    Question & Rephrased Doctor's Response \\
    \hline\hline
    Is my Anti HIV Test Conclusive after 241 days or need retest? & The time frame and final results of the test indicate that the patient is HIV-free even after 241 days.\\
    \hline
    I developed breathing difficulties approx. 2 years ago, which have worsened daily since. What can I do? & In my opinion, if there is something other than anxiety, then the possibility of thyroid and hiatal hernia of the esophagus should be checked.\\
    \hline
    Low resting heart rate. Can it be secondary adrenal insufficiency? If not, what is the cause? & This is normal after a stressful and traumatic event and nothing to worry about.\\
    \hline\hline
  \end{tabular}
\end{table*}

\section{Corpus Analysis}
The dataset consists of 1880 distinct texts each for the doctor's response, the ChatGPT response, and the rephrased doctor's response, resulting in a total of 5640 text inputs. The dataset mainly contains short texts of 250 characters or fewer, as shown in Figure~\ref{fig:fig5}. The dataset was divided into three separate files for three different experiments, as presented in~\ref{tab:tab7}. The first file contains responses obtained from both the doctor and the ChatGPT model. The second file consists of responses from the doctor, along with rephrased versions of the doctor's responses. The third file contains the doctor, ChatGPT, and the rephrased doctor's responses, which were combined for the multi-label classification task. 

\begin{figure*}[!ht]
  \centering
  \includegraphics[width=0.8\textwidth]{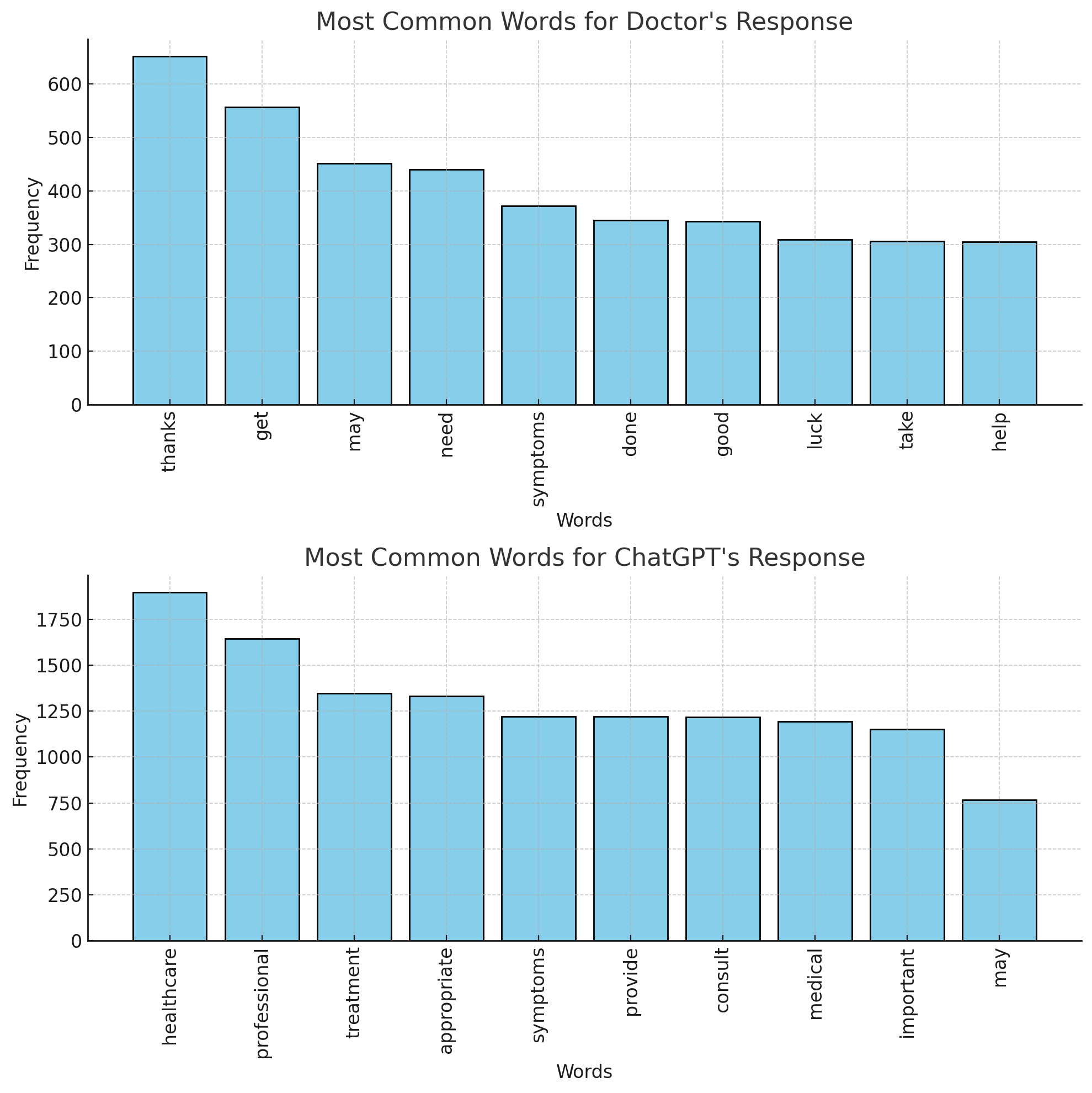}
  \caption{Most commonly used words per label.}
  \label{fig:fig1}
\end{figure*}

\begin{figure*}[!ht]
  \centering
  \includegraphics[width=0.8\textwidth]{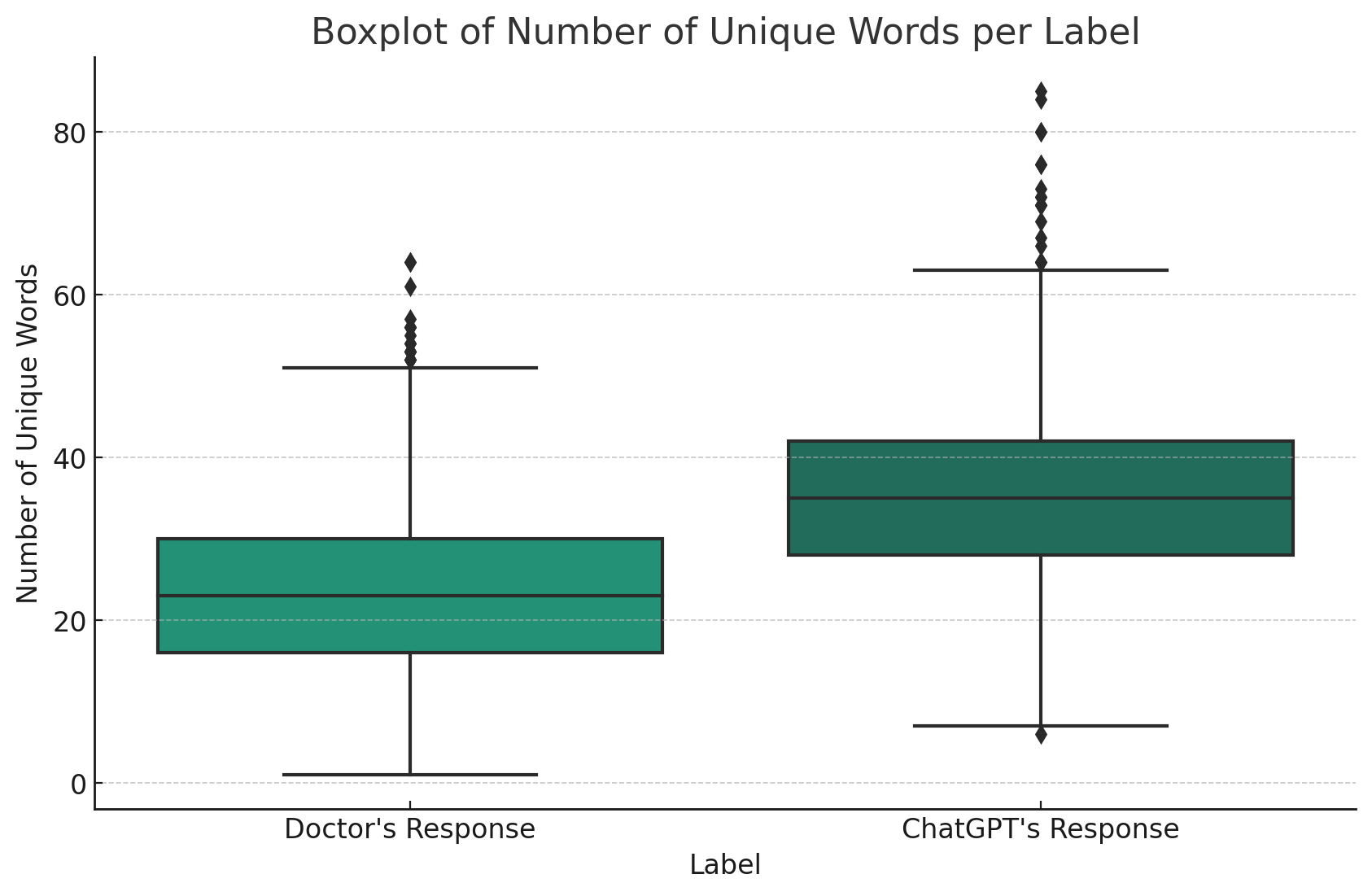}
  \caption{Number of unique words per label.}
  \label{fig:fig2}
\end{figure*}

\begin{figure*}[!ht]
  \centering
  \includegraphics[width=0.8\textwidth]{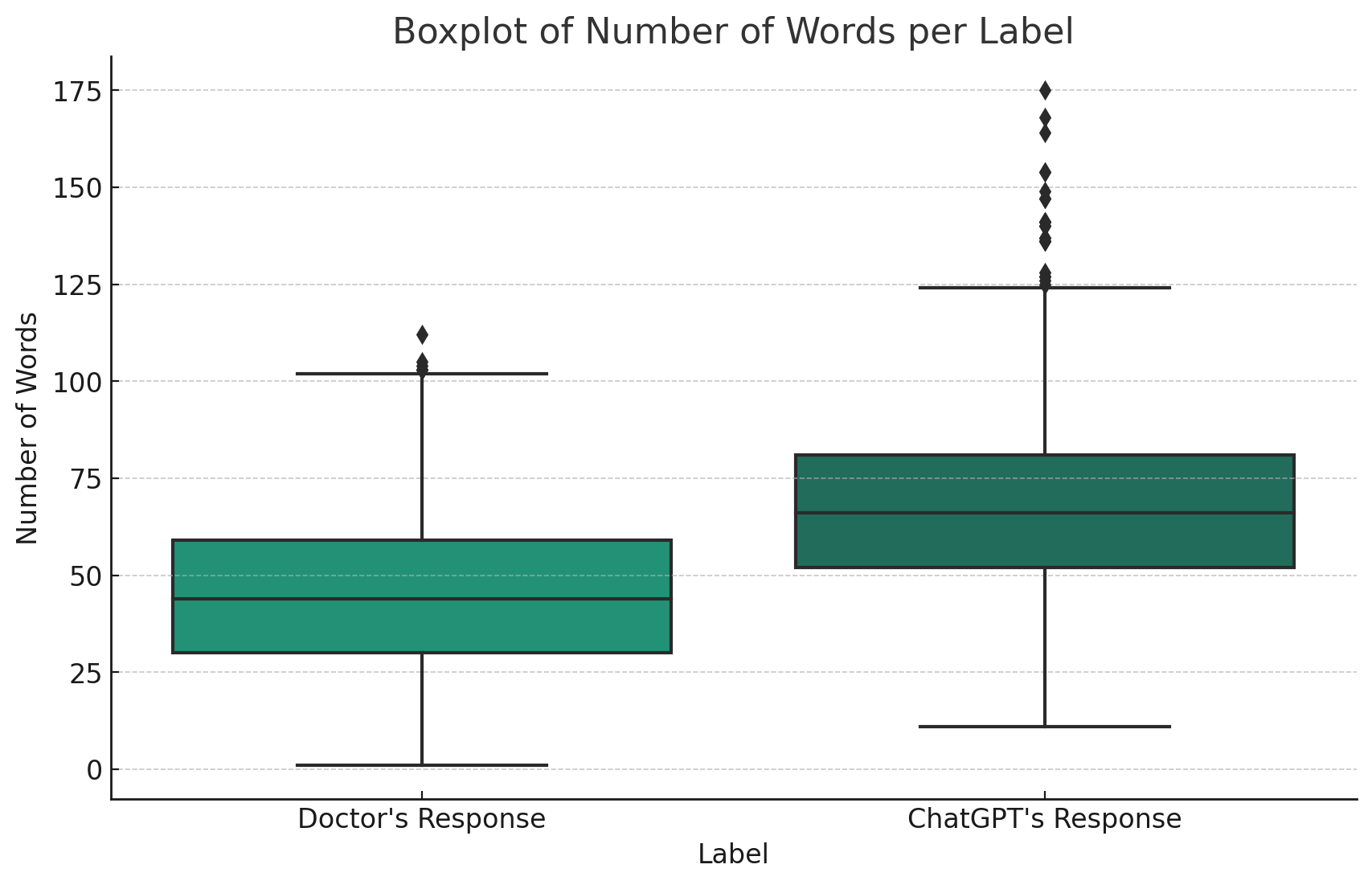}
  \caption{Number of words per label.}
  \label{fig:fig3}
\end{figure*}

\begin{figure*}[!ht]
  \centering
  \includegraphics[width=0.8\textwidth]{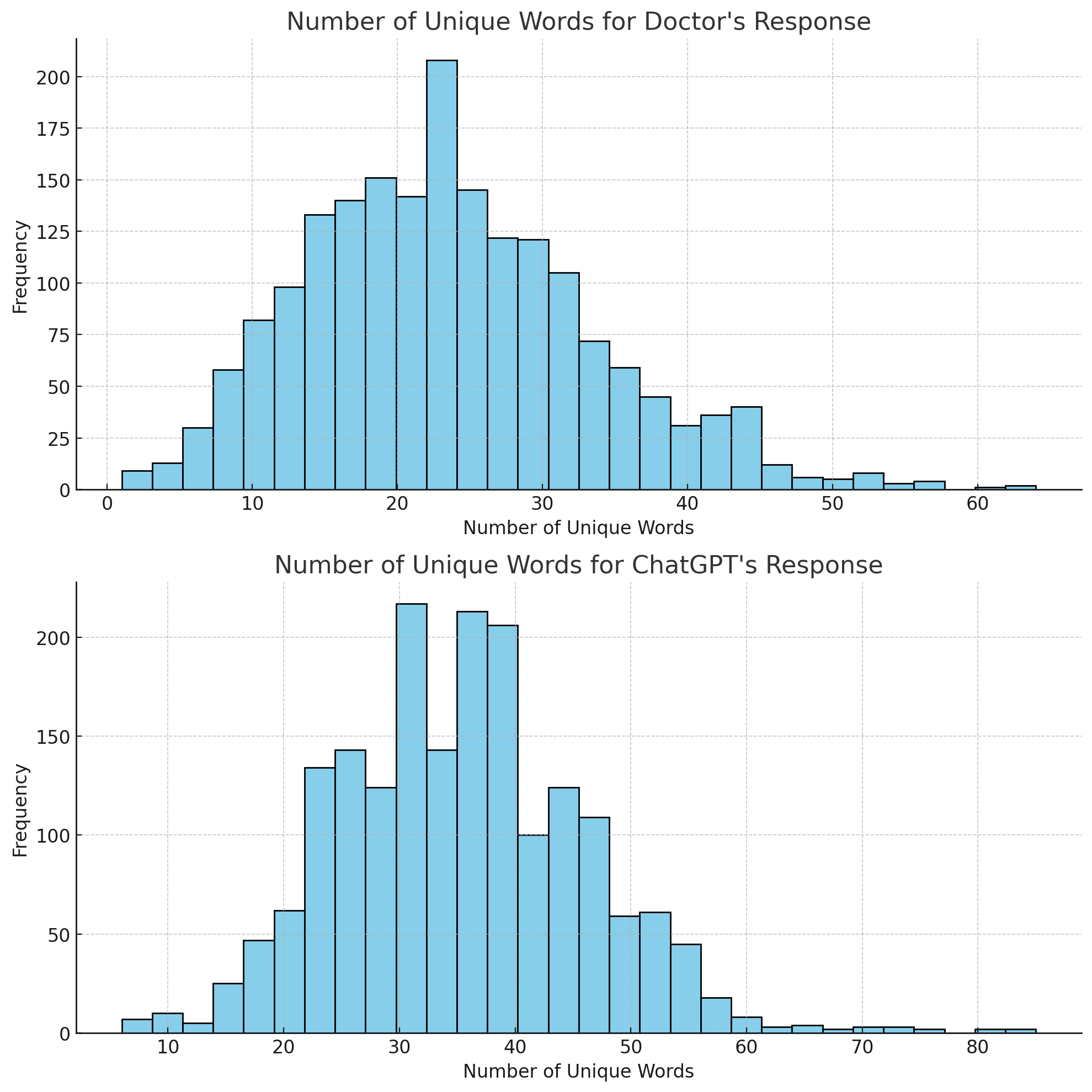}
  \caption{Number of unique words per label.}
  \label{fig:fig4}
\end{figure*}

\begin{figure*}[!ht]
  \centering
  \includegraphics[width=0.8\textwidth]{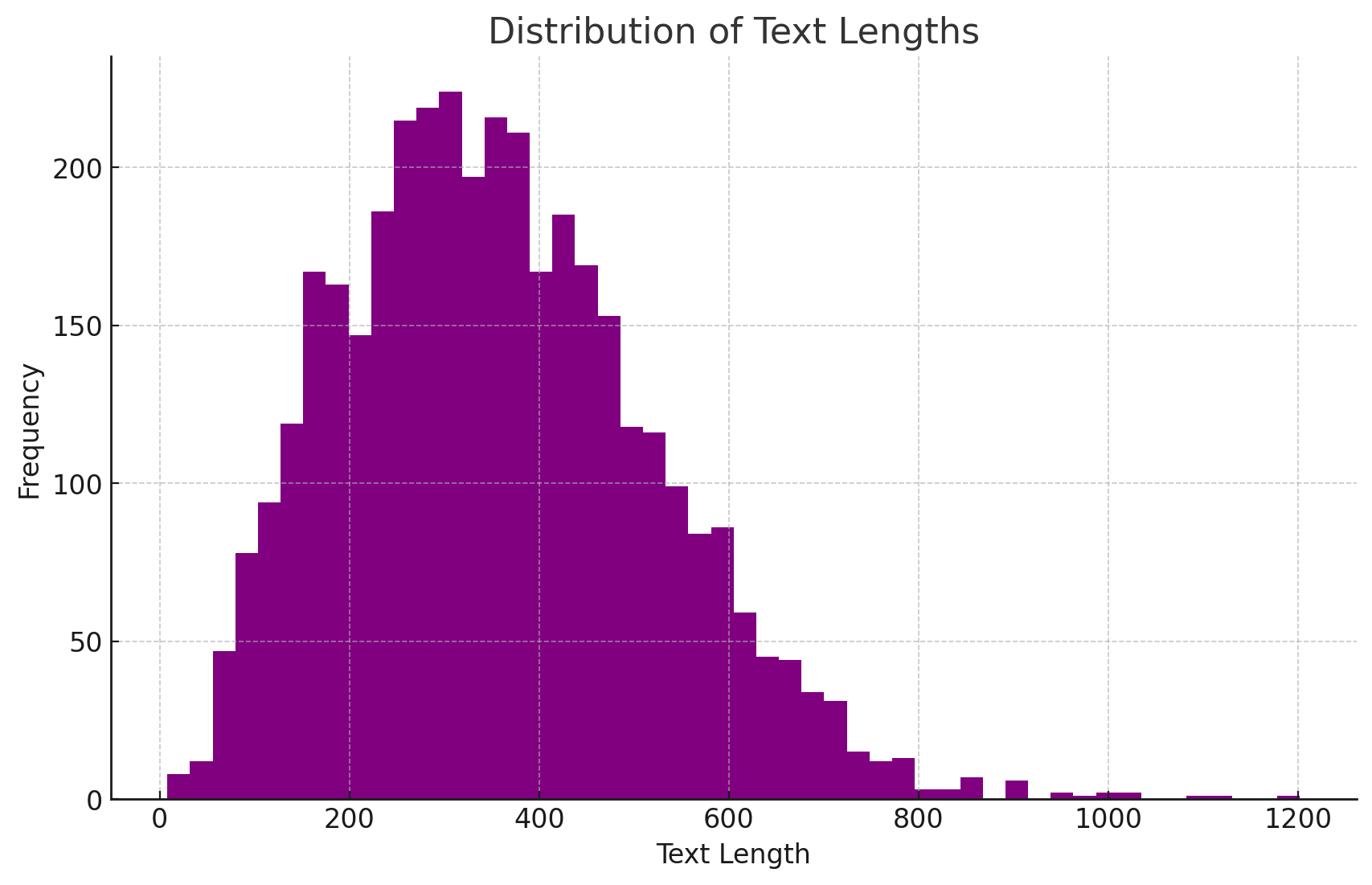}
  \caption{Distribution of text lengths in the combined dataset.}
  \label{fig:fig5}
\end{figure*}

We analyzed the DC dataset as shown in figures~\ref{fig:fig1}-~\ref{fig:fig4}. Since the rephrased response is a modified version of the doctor's response, we expect it to produce a similar result. As the dataset contains answers to questions from both doctors and models, it is a balanced representation that will not be biased towards any class. Examination of text lengths and the number of words per response indicates a higher median text length for the ChatGPT responses compared to doctor responses (Figure~\ref{fig:fig2}). This could suggest that ChatGPT tends to provide more words or complex responses than doctors. In figure~\ref{fig:fig3}, the boxplot of the number of words per response reveals a similar pattern, indicating that ChatGPT's responses generally contain more words.

Looking further into the content of the responses, we cleaned the text data by removing stopwords and punctuations and then extracted the most common words for each response category. The bar graphs in figure~\ref{fig:fig1} display the top ten most common words after the cleaning process for each category. These most common words provide a glimpse into the topics or themes present in the responses. For instance, words like 'healthcare', 'professional', 'treatment', and 'appropriate' frequently occur in the ChatGPT responses, indicating a clinical focus in their communication. Several texts generated by ChatGPT emphasized the importance of seeking medical advice from a licensed professional or visiting a hospital. We also plotted the unique words per response (figure~\ref{fig:fig4}), showing that the responses of the doctor and ChatGPT models exhibit a wide range of unique word counts, indicating a diverse vocabulary in the corpus. However, doctors' responses tend to have a slightly higher number of unique words, which might suggest a broader range of topics or more complex discussions in their communication.

\section{Experimental Analysis}
In a supervised learning approach, a model can be taught how to map new and unseen inputs to correct outputs. By employing transfer learning, zero-shot classification models achieve precise classification results for data from classes unseen in the training phase. This principle lies in the model's capacity to capture the semantic information of the words or sentences and learn from patterns that were not present in the training data. In this research, we examine whether zero-shot transformer models can learn patterns not represented in the training data. This is a challenging task, as ML models typically struggle to generalize to patterns they have not seen during training. The diagram showing the experimental setup is illustrated in figure~\ref{fig:fig6}.

To determine whether pre-trained language models are effective in classifying doctor- and AI-generated responses in health consultations without corpus training, we performed a series of experiments using zero-shot classification algorithms. The results of different zero-shot classification models for classifying ChatGPT, doctor-generated and rephrased doctor responses in healthcare consultations are presented in Tables~\ref{tab:tab4}-~\ref{tab:tab6}. The evaluation metrics used are accuracy, precision, recall, and the F1 score. The models evaluated in this study are the Bidirectional Auto-Regressive Transformers (BART)~\cite{lewis2019bart}, Bidirectional Encoder Representations from Transformers (BERT)~\cite{devlin2018bert}, Cross-lingual Language Model (XLM)~\cite{lample2019cross}, Cross-lingual Language Model - RoBERTa (XLM-R)~\cite{conneau2019unsupervised}, and Distill-Bidirectional Encoder Representations from Transformers (DistilBERT)~\cite{sanh2019distilbert}. These models were tested on the three different datasets in a binary- and multi-label classification tasks.

The analysis revealed that all models struggled to accurately classify the datasets when training was not performed. These findings suggest that additional training of models using the corpus may be required to improve their classification performance. According to Table~\ref{tab:tab4}, BART performs the best when classifying the DC dataset, followed by XLM, according to F1 scores. Table~\ref{tab:tab5} shows moderate classification results when classifying the DR dataset, with the BART model outperforming others. The analysis performed in Table~\ref{tab:tab6} reveals a notable decrease in the performance of the models when classifying the DCR dataset, with BERT having the highest F1 score.

\begin{table*}[!ht]
  \centering
  \small
  \caption{Performance Evaluation of Zero-shot Classification Models for the DC Dataset}
  \label{tab:tab4}
  \begin{tabular}{c|c|c|c|c} 
    \hline\hline
    Model & Accuracy & Precision & Recall & F1 \\
    \hline\hline
    BART & 0.5851 & 0.5872 & 0.5851 & 0.5825 \\
    \hline
    BERT & 0.4649 & 0.4602 & 0.4649 & 0.4486 \\
    \hline
    XLM & 0.5037 & 0.5038 & 0.5037 & 0.5022 \\
    \hline
    XLM-R & 0.5314 & 0.7350 & 0.5314 & 0.4018 \\
    \hline
    DistilBERT & 0.4625 & 0.4622 & 0.4625 & 0.4616 \\
    \hline\hline
  \end{tabular}
\end{table*}

\begin{table*}[!ht]
  \centering
  \small
  \caption{Performance Evaluation of Zero-shot Classification Models for the DR Dataset}
  \label{tab:tab5}
  \begin{tabular}{c|c|c|c|c}  
    \hline\hline
    Model & Accuracy & Precision & Recall & F1 \\
    \hline\hline
    BART & 0.5223 & 0.5224 & 0.5223 & 0.5222 \\
    \hline
    BERT & 0.4822 & 0.4804 & 0.4822 & 0.4701 \\
    \hline
    XLM & 0.5122 & 0.5156 & 0.5122 & 0.4843 \\
    \hline
    XLM-R & 0.4997 & 0.4867 & 0.4997 & 0.3374 \\
    \hline
    DistilBERT & 0.4801 & 0.4792 & 0.4801 & 0.4748 \\
    \hline\hline
  \end{tabular}
\end{table*}

\begin{table*}[!ht]
  \centering
  \small
  \caption{Performance Evaluation of Zero-shot Classification Models for the DCR Dataset}
  \label{tab:tab6}
  \begin{tabular}{c|c|c|c|c} 
    \hline\hline
    Model & Accuracy & Precision & Recall & F1 \\
    \hline\hline
    BART & 0.3913 & 0.3838 & 0.3913 & 0.3434 \\
    \hline
    BERT & 0.4806 & 0.5158 & 0.4806 & 0.4261 \\
    \hline
    XLM & 0.3262 & 0.3306 & 0.3262 & 0.2857 \\
    \hline
    XLM-R & 0.3239 & 0.3249 & 0.3239 & 0.2263 \\
    \hline
    DistilBERT & 0.3344 & 0.2851 & 0.3344 & 0.1853 \\
    \hline\hline
  \end{tabular}
\end{table*}

\section{Discussion}
We address the challenging task of zero-shot classification in the context of health consultations, specifically focusing on distinguishing between responses generated by Doctors and those generated by AI systems. The increasing integration of AI technologies in healthcare has led to the coexistence of human expertise and machine-generated insights in medical consultations. As such, it becomes imperative to develop robust methods that can discern and classify responses originating from these distinct sources. The results of our study demonstrate the strengths and limitations of various zero-shot classification models in the classification of health consultations. Despite the challenges encountered, the potential of transfer learning and zero-shot classification remains promising, and with further fine-tuning and specific corpus training, these models could prove to be valuable tools in improving medical consultations and decision-making processes. In the three different sets of datasets used to evaluate the zero-shot classification models, each model's performance varied across the datasets, highlighting its strengths and weaknesses. BART achieved, on average, the highest F1 score in the three experiments. The performance of all models decreased when applied to the combined set of responses, indicating the challenge posed by the increased complexity and diversity of the data. These limitations in model performance raise concerns about their reliability and accuracy when applied to healthcare consultations. Considering that the models' writing style and patterns keep changing, not being able to classify them without specific training could lead to wrong classifications. This could be a serious problem in healthcare because we will not have enough data to train the models properly when their writing style changes. In our future work, we plan to develop more accurate and robust text classification models for health consultations by using few-shot learning with different embedding techniques. 

\section{Conclusion}
In this research, we explored the effectiveness of zero-shot classification models in classifying Doctor- and AI-generated responses in health consultations without corpus training. Our research contributes to understanding the capabilities of zero-shot classification models and their applicability in the classification of Doctor-generated and AI-generated medical texts. We developed zero-shot classification models to distinguish between responses provided by Doctors and those provided by AI systems in healthcare consultations. Based on the performance evaluation of the zero-shot classification models on the different datasets, we conclude that these models may not be the optimal choice for the classification of Doctor-generated and AI-generated text. The decrease in model performance on the combined multi-labeled text underscores the difficulty of accurately classifying a diverse set of the AI-generated text. This might be due to the complex sentence structures often found in such texts, which may require a deeper understanding of context or specialized knowledge than what these models currently offer. Our future work will explore the application of specialized, supervised, and few-shot learning classification techniques to handle the complexity and diversity of this type of classification task. Although zero-shot classification models may perform well for some other text classification tasks, they may not be the best fit for tasks involving doctor- and AI-generated texts.

\section{Acknowledgments}
This work was done with partial support from the Mexican Government through the grant A1-S-47854 of CONACYT, Mexico, grants 20232138, 20230140, 20232080 and 20231567 of the Secretaría de Investigación y Posgrado of the Instituto Politécnico Nacional, Mexico. The authors thank CONACYT for the computing resources brought to them through the Plataforma de Aprendizaje Profundo para Tecnologías del Lenguaje of the Laboratorio de Supercómputo of the INAOE, Mexico and acknowledge the support of Microsoft through the Microsoft Latin America PhD Award.

\bibliographystyle{arxiv}
\bibliography{references}

\begin{thebibliography}{10}
\expandafter\ifx\csname url\endcsname\relax
  \def\url#1{\texttt{#1}}\fi
\expandafter\ifx\csname urlprefix\endcsname\relax\def\urlprefix{URL }\fi
\expandafter\ifx\csname href\endcsname\relax
  \def\href#1#2{#2} \def\path#1{#1}\fi

\bibitem{snelling2014introduction}
A.~M. Snelling, Introduction to health promotion, John Wiley \& Sons, 2014.

\bibitem{mitchell2008multidisciplinary}
G.~K. Mitchell, J.~J. Tieman, T.~M. Shelby-James, Multidisciplinary care planning and teamwork in primary care, Medical Journal of Australia 188 (2008) S61--S64.

\bibitem{dal2023challenges}
F.~Dal~Mas, M.~Massaro, P.~Rippa, G.~Secundo, The challenges of digital transformation in healthcare: An interdisciplinary literature review, framework, and future research agenda, Technovation 123 (2023) 102716.

\bibitem{siala2022shifting}
H.~Siala, Y.~Wang, Shifting artificial intelligence to be responsible in healthcare: A systematic review, Social Science \& Medicine 296 (2022) 114782.

\bibitem{adams2017dehumanizing}
S.~M. Adams, T.~I. Case, J.~Fitness, R.~J. Stevenson, Dehumanizing but competent: The impact of gender, illness type, and emotional expressiveness on patient perceptions of doctors, Journal of Applied Social Psychology 47~(5) (2017) 247--255.

\bibitem{yun2021behavioral}
J.~H. Yun, E.-J. Lee, D.~H. Kim, Behavioral and neural evidence on consumer responses to human doctors and medical artificial intelligence, Psychology \& Marketing 38~(4) (2021) 610--625.

\bibitem{ojo2022language}
O.~Ojo, A.~Gelbukh, H.~Calvo, A.~Feldman, O.~Adebanji, J.~Armenta-Segura, Language identification at the word level in code-mixed texts using character sequence and word embedding, in: Proceedings of the 19th International Conference on Natural Language Processing (ICON): Shared Task on Word Level Language Identification in Code-mixed Kannada-English Texts, 2022, pp. 1--6.

\bibitem{adebanji2022sequential}
O.~O. Adebanji, I.~Gelbukh, H.~Calvo, O.~E. Ojo, Sequential models for sentiment analysis: A comparative study, in: Mexican International Conference on Artificial Intelligence, Springer, 2022, pp. 227--235.

\bibitem{9677627}
A.~E. Blanchard, S.~Gao, H.-J. Yoon, J.~B. Christian, E.~B. Durbin, X.-C. Wu, A.~Stroup, J.~Doherty, S.~M. Schwartz, C.~Wiggins, L.~Coyle, L.~Penberthy, G.~D. Tourassi, A keyword-enhanced approach to handle class imbalance in clinical text classification, IEEE Journal of Biomedical and Health Informatics 26~(6) (2022) 2796--2803.
\newblock \href {http://dx.doi.org/10.1109/JBHI.2022.3141976} {\path{doi:10.1109/JBHI.2022.3141976}}.

\bibitem{prabhakar2021medical}
S.~K. Prabhakar, D.-O. Won, Medical text classification using hybrid deep learning models with multihead attention, Computational intelligence and neuroscience 2021.

\bibitem{9451752}
P.~Lavanya, E.~Sasikala, Deep learning techniques on text classification using natural language processing ({NLP}) in social healthcare network: A comprehensive survey, in: 2021 3rd International Conference on Signal Processing and Communication (ICPSC), 2021, pp. 603--609.
\newblock \href {http://dx.doi.org/10.1109/ICSPC51351.2021.9451752} {\path{doi:10.1109/ICSPC51351.2021.9451752}}.

\bibitem{kumar2022deep}
Y.~Kumar, A.~Koul, S.~Mahajan, A deep learning approaches and fast{AI} text classification to predict 25 medical diseases from medical speech utterances, transcription and intent, Soft computing 26~(17) (2022) 8253--8272.

\bibitem{LI2021345}
X.~Li, M.~Cui, J.~Li, R.~Bai, Z.~Lu, U.~Aickelin, \href{https://www.sciencedirect.com/science/\\article/pii/S0925231221003258}{A hybrid medical text classification framework: Integrating attentive rule construction and neural network}, Neurocomputing 443 (2021) 345--355.
\newblock \href {http://dx.doi.org/https://doi.org/10.1016/j.neucom.2021.02.069} {\path{doi:https://doi.org/10.1016/j.neucom.2021.02.069}}.
\newline\urlprefix\url{https://www.sciencedirect.com/science/\\article/pii/S0925231221003258}

\bibitem{liao2023differentiate}
W.~Liao, Z.~Liu, H.~Dai, S.~Xu, Z.~Wu, Y.~Zhang, X.~Huang, D.~Zhu, H.~Cai, T.~Liu, et~al., Differentiate chat{GPT}-generated and human-written medical texts, arXiv preprint arXiv:2304.11567.

\bibitem{ojo2023legend}
O.~E. Ojo, O.~O. Adebanji, H.~Calvo, D.~O. Dieke, O.~E. Ojo, S.~E. Akinsanya, T.~O. Abiola, A.~Feldman, Legend at araieval shared task: Persuasion technique detection using a language-agnostic text representation model (2023).
\newblock \href {http://arxiv.org/abs/2310.09661} {\path{arXiv:2310.09661}}.

\bibitem{ojo2020sentiment}
O.~E. Ojo, A.~Gelbukh, H.~Calvo, O.~O. Adebanji, G.~Sidorov, Sentiment detection in economics texts, in: Mexican International Conference on Artificial Intelligence, Springer, 2020, pp. 271--281.

\bibitem{ojo2021performance}
O.~Ojo, A.~Gelbukh, H.~Calvo, O.~Adebanji, Performance study of n-grams in the analysis of sentiments, Journal of the Nigerian Society of Physical Sciences (2021) 477--483.

\bibitem{ojo2023transformer}
O.~E. Ojo, H.~T. Ta, A.~Gelbukh, H.~Calvo, O.~O. Adebanji, G.~Sidorov, Transformer-based approaches to sentiment detection, in: Recent Developments and the New Directions of Research, Foundations, and Applications: Selected Papers of the 8th World Conference on Soft Computing, February 03--05, 2022, Baku, Azerbaijan, Vol. II, Springer, 2023, pp. 101--110.

\bibitem{srivastava2020healthcare}
S.~K. Srivastava, S.~K. Singh, J.~S. Suri, A healthcare text classification system and its performance evaluation: A source of better intelligence by characterizing healthcare text, in: Cognitive informatics, computer modelling, and cognitive science, Elsevier, 2020, pp. 319--369.

\bibitem{ojo2022automatic}
O.~E. Ojo, T.-H. Ta, A.~Gelbukh, H.~Calvo, G.~Sidorov, O.~O. Adebanji, Automatic hate speech detection using deep neural networks and word embedding, Computaci{\'o}n y Sistemas 26~(2) (2022) 1007--1013.

\bibitem{9185650}
C.~Tu, M.~Cui, Learning regular expressions for interpretable medical text classification using a pool-based simulated annealing approach, in: 2020 IEEE Congress on Evolutionary Computation (CEC), 2020, pp. 1--7.
\newblock \href {http://dx.doi.org/10.1109/CEC48606.2020.9185650} {\path{doi:10.1109/CEC48606.2020.9185650}}.

\bibitem{mitrovic2023chatgpt}
S.~Mitrovi{\'c}, D.~Andreoletti, O.~Ayoub, Chatgpt or human? detect and explain. explaining decisions of machine learning model for detecting short chat{GPT}-generated text, arXiv preprint arXiv:2301.13852.

\bibitem{Faqar-Uz-Zamane041396}
S.~F. Faqar-Uz-Zaman, N.~Filmann, D.~Mahkovic, M.~von Wagner, C.~Detemble, U.~Kippke, U.~Marschall, L.~Anantharajah, P.~Baumartz, P.~Sobotta, W.~O. Bechstein, A.~A. Schnitzbauer, \href{https://bmjopen.bmj.com/content/11/1/\\e041396}{Study protocol for a prospective, double-blinded, observational study investigating the diagnostic accuracy of an app-based diagnostic health care application in an emergency room setting: the eradar trial}, BMJ Open 11~(1).
\newblock \href {http://arxiv.org/abs/https://bmjopen.bmj.com/content/11/1/e041396.full.pdf} {\path{arXiv:https://bmjopen.bmj.com/content/11/1/e041396.full.pdf}}, \href {http://dx.doi.org/10.1136/bmjopen-2020-041396} {\path{doi:10.1136/bmjopen-2020-041396}}.
\newline\urlprefix\url{https://bmjopen.bmj.com/content/11/1/\\e041396}

\bibitem{melvin-etal-2004-creation}
R.~S. Melvin, W.~May, S.~Narayanan, P.~Georgiou, S.~Ganjavi, \href{http://www.lrec-conf.org/proceedings/\\lrec2004/pdf/391.pdf}{Creation of a doctor-patient dialogue corpus using standardized patients}, in: Proceedings of the Fourth International Conference on Language Resources and Evaluation ({LREC}{'}04), European Language Resources Association (ELRA), Lisbon, Portugal, 2004, pp. 1--4.
\newline\urlprefix\url{http://www.lrec-conf.org/proceedings/\\lrec2004/pdf/391.pdf}

\bibitem{veeranna2016using}
S.~P. Veeranna, J.~Nam, E.~L. Menc{\i}a, J.~F{\"u}rnkranz, Using semantic similarity for multi-label zero-shot classification of text documents, in: Proceeding of European Symposium on Artificial Neural Networks, Computational Intelligence and Machine Learning. Bruges, Belgium: Elsevier, 2016, pp. 423--428.

\bibitem{baker2020comparison}
A.~Baker, Y.~Perov, K.~Middleton, J.~Baxter, D.~Mullarkey, D.~Sangar, M.~Butt, A.~DoRosario, S.~Johri, A comparison of artificial intelligence and human doctors for the purpose of triage and diagnosis, Frontiers in artificial intelligence 3 (2020) 543405.

\bibitem{shoham2023cpllm}
O.~B. Shoham, N.~Rappoport, Cpllm: Clinical prediction with large language models, arXiv preprint arXiv:2309.11295.

\bibitem{dave2023chatgpt}
T.~Dave, S.~A. Athaluri, S.~Singh, {ChatGPT} in medicine: an overview of its applications, advantages, limitations, future prospects, and ethical considerations, Frontiers in Artificial Intelligence 6 (2023) 1169595.

\bibitem{lupart2022zero}
S.~Lupart, B.~Favre, V.~Nikoulina, S.~Ait-Mokhtar, Zero-shot and few-shot classification of biomedical articles in context of the covid-19 pandemic, arXiv preprint arXiv:2201.03017.

\bibitem{mascio2020comparative}
A.~Mascio, Z.~Kraljevic, D.~Bean, R.~Dobson, R.~Stewart, R.~Bendayan, A.~Roberts, Comparative analysis of text classification approaches in electronic health records, arXiv preprint arXiv:2005.06624.

\bibitem{liu2023deid}
Z.~Liu, X.~Yu, L.~Zhang, Z.~Wu, C.~Cao, H.~Dai, L.~Zhao, W.~Liu, D.~Shen, Q.~Li, et~al., Deid-{GPT}: Zero-shot medical text de-identification by {GPT}-4, arXiv preprint arXiv:2303.11032.

\bibitem{8733799}
K.~Liu, L.~Chen, Medical social media text classification integrating consumer health terminology, IEEE Access 7 (2019) 78185--78193.
\newblock \href {http://dx.doi.org/10.1109/ACCESS.2019.2921938} {\path{doi:10.1109/ACCESS.2019.2921938}}.

\bibitem{8991726}
H.~Jiang, C.~Chen, S.~Wu, Y.~Guo, Classification of medical sensitive data based on text classification, in: 2019 IEEE International Conference on Consumer Electronics - Taiwan (ICCE-TW), 2019, pp. 1--2.
\newblock \href {http://dx.doi.org/10.1109/ICCE-TW46550.2019.8991726} {\path{doi:10.1109/ICCE-TW46550.2019.8991726}}.

\bibitem{QuestionDoctors}
Questiondoctors, \url{https://questiondoctors.com/}.

\bibitem{github}
Github, \url{https://github.com/LasseRegin/medical-\\question-answer-data/blob/master/README.md}.

\bibitem{chatgpt}
Chat{GPT}, \url{https://chat.openai.com/}.

\bibitem{rapid}
R.~API, \href{https://rapidapi.com/neuralwriter-\\neuralwriter-default/api/paraphrasing-and-rewriter-api/}{Paraphrasing and rewriter api} (2023).
\newline\urlprefix\url{https://rapidapi.com/neuralwriter-\\neuralwriter-default/api/paraphrasing-and-rewriter-api/}

\bibitem{lewis2019bart}
M.~Lewis, Y.~Liu, N.~Goyal, M.~Ghazvininejad, A.~Mohamed, O.~Levy, V.~Stoyanov, L.~Zettlemoyer, {BART}: Denoising sequence-to-sequence pre-training for natural language generation, translation, and comprehension, arXiv preprint arXiv:1910.13461.

\bibitem{devlin2018bert}
J.~Devlin, M.-W. Chang, K.~Lee, K.~Toutanova, {BERT}: Pre-training of deep bidirectional transformers for language understanding, arXiv preprint arXiv:1810.04805.

\bibitem{lample2019cross}
G.~Lample, A.~Conneau, Cross-lingual language model pretraining, arXiv preprint arXiv:1901.07291.

\bibitem{conneau2019unsupervised}
A.~Conneau, K.~Khandelwal, N.~Goyal, V.~Chaudhary, G.~Wenzek, F.~Guzm{\'a}n, E.~Grave, M.~Ott, L.~Zettlemoyer, V.~Stoyanov, Unsupervised cross-lingual representation learning at scale, arXiv preprint arXiv:1911.02116.

\bibitem{sanh2019distilbert}
V.~Sanh, L.~Debut, J.~Chaumond, T.~Wolf, {DistilBERT}, a distilled version of {BERT}: smaller, faster, cheaper and lighter, arXiv preprint arXiv:1910.01108.

\end{thebibliography}

\end{document}